# Dynamic Feature Scaling for K-Nearest Neighbor Algorithm

Chandrasekaran Anirudh Bhardwaj[1], Megha Mishra [1], Kalyani Desikan[2]

[1]*B-Tech, School of computing science and engineering, VIT University, Chennai, India*
[1]*B-Tech, School of computing science and engineering, VIT University, Chennai, India*
[2]*Faculty-School of Advanced Sciences, VIT University, Chennai, India*

*E-mail:*[1] anirudhsekar96@gmail.com, [1]meghamishra04@gmail.com, [2]kalyanidesikan@vit.ac.in

## Abstract

Nearest Neighbors Algorithm is a Lazy Learning Algorithm, in which the algorithm tries to approximate the predictions with the help of similar existing vectors in the training dataset. The predictions made by the K-Nearest Neighbors algorithm is based on averaging the target values of the spatial neighbors. The selection process for neighbors in the Hermitian space is done with the help of distance metrics such as Euclidean distance, Minkowski distance, Mahalanobis distance etc. A majority of the metrics such as Euclidean distance are scale variant, meaning that the results could vary for different range of values used for the features. Standard techniques used for the normalization of scaling factors are feature scaling method such as Z-score normalization technique, Min-Max scaling etc. Scaling methods uniformly assign equal weights to all the features, which might result in a non-ideal situation. This paper proposes a novel method to assign weights to individual feature with the help of out of bag errors obtained from constructing multiple decision tree models.

## Introduction

Nearest Neighbors Algorithm is one of the oldest and most studied algorithms used in the history of learning algorithms. It uses an approximating function, which averages the target values for the unknown vector with the help of target values of the vectors which are a part of the training dataset which are closest to the unknown vector in the Hermitian Space. K-Nearest Neighbors algorithm[1] takes Nearest Neighbors Algorithm a step forward by taking into account the target values of the K-nearest vectors from the training dataset. K-Nearest Neighbors is used for Classification as well as for Regression. K-Nearest is widely used in multiple areas like Text Categorization [2], Recommender systems, Credit Scoring [3], Handwriting Recognition, Speech Recognition, Predicting Economic Events [4] etc.

All the neighbors are considered equal in standard K-Nearest Neighbors i.e all the data points have equal votes [5] in determining the class of their neighboring target data point. However, their contribution may vary based on their distance from the target. The data points are highly dependent on their neighbors. The class of Nearest Neighbors Algorithms, which includes K-Nearest Neighbors Algorithm (henceforth abbreviated as KNN), utilize distance metrics to identify the closest neighbors. There are multiple distance metrics which can be used such as

Manhattan distance, Euclidean distance, Minkowski distance etc. In the K-Nearest Neighbors algorithm distance metrics such as Euclidean distance[6], Mahalanobis distance[6], City-block[6] distance, Minkowski[6] distance, Chebychev[6] distance and Standardized Euclidean distance metrics are considered as some of the best metrics for providing the highest accuracy [7] for classification as well as regression tasks.

Most of the metrics used are scale variant, meaning that the values for the distances between any two points change drastically with a change in the scale of values used in a particular dimension. Distance metrics such as Mahalanobis distance try to uniformly scale the dimensions to assign each feature a uniform weight. Methods for scaling such as Z-score normalization and Min-Max scaling are used to assign uniform weights to features for scale variant distance metrics.

Assigning uniform feature weights is generally a good strategy to increase the accuracy of the predictions made by spatial algorithms such as KNN, but is not optimal strategy in all the cases. In cases where several noisy features are present in the dataset, assigning uniform weights to all the features might result in poor accuracy of prediction. Noisy features [8] refer to the attributes that do not have any accountable contribution to the predictions made, and might contain data prone to more erroneous results.

Feature weighing takes into account the relevance of the features used, and assigns weights to them. Most of the methods used for assigning feature weights have to train and retrain the KNN model for getting a good enough result. This results in a huge performance overhead.

An approach which does not require the need of repeatedly training the KNN model to achieve a higher accuracy is needed. This can be solved with the help of assigning feature weights based on the feature importance obtained while training a Random Forest Model. Random Forest model constructs many decision trees with the help of Random Subspace method. Random Subspace method further injects randomization for increasing the diversity of the models used for ensembling by randomly selecting the feature subsets for training the learning algorithm in excess to choosing a subset of all the samples to train with the help of Bootstrap sampling.

Constructing different models on different subsets of dimensions leads to an easy calculation of out-of-bag error for each particular dimension, which can further be used as a metric for feature importance.

The proposed method, called Feature Importance based Dynamic Scaling method (henceforth abbreviated as FIDS), uses feature importance value gained from training a Random Forest algorithm to scale features dynamically to increase the accuracy of the predictions.

## Literature Review

Nearest Neighbors algorithm, and its variations such as the K-Nearest Neighbors algorithm, belong to a class of lazy learners. Lazy learning algorithms compute the results as and when the data to be predicted becomes available. These algorithms have a benefit of avoiding the phenomenon of co-variate shift[9] due to change in the data patterns as time progresses. This means that algorithms like KNN take less time to train, and adapt to changes in patterns in the data over time. However, the test time is equivalently high as compared to the training time partly because while training only the number of neighborhood points i.e. K value of the K Nearest Neighbors Algorithm needs to be chosen whereas during testing the distance of each point in the test set needs to be computed versus all the points in the training set.

KNN depends explicitly on the training dataset [10] since it only recalls all the classifications made in training dataset and does not find any particular pattern for identifying the class of the new data point. Therefore, whenever a new test data is introduced, KNN has to go through all the train data points to find a perfect match or at the very least an approximate match. This increases the overall time needed for testing the algorithm on previously unseen data. This also means that the K-Nearest Neighbors algorithm requires a larger space to store the complete training dataset.

Another problem that arises with the K-Nearest Neighbors algorithm is the phenomenon of imbalanced class problem [11,12,13]. The KNN algorithm tends to give a biased [14] result in the favor of larger classes, meaning there is a higher probability [14] for a new data point to be classified in the class having larger size. This problem is known popularly as the imbalanced class problem.

The KNN algorithm uses metrics such as Euclidean distance[6][7], Manhattan distance[6][7], Minkowski distance[6][7] etc. to compute the distance between different points and identify the neighbors for the various samples that are to be used to predict the target values. These metrics are influenced by the range of values on which each feature value is based on, meaning that a feature with a greater range of values for scaling would hold more importance than a feature with a smaller range of values. The process of increasing the range of values for a particular feature, and thereby increasing the importance of the particular feature is called stretching, and correspondingly decreasing the range of values for a particular feature is called shrinking. Generally, the initial weights are assigned uniformly with the help of scaling techniques such as Min-Max Scaling and Z-Score normalization[15]. Though this may result in a better result as compared to an unscaled variant, this is generally not the optimal strategy.

Accuracy of the predictions made can be increased by utilizing strategies such as dimensionality reduction, feature selection, feature extraction, feature weighing, metric learning and hyper-parameter optimization. Dimensionality reduction techniques such as Principal Component Analysis[16] help in reducing the redundancy in the data by reorienting and projecting the results in a space with different basis. Feature Extraction is the process of extracting relevant features and incorporating domain knowledge into the prediction algorithms to further increase the accuracy of the predictions. Feature selection methods such as Stepwise Forward Selection[17], Stepwise Backward Elimination[17] and Genetic algorithms[18] tries to find relevant features necessary, and tend to drop the unnecessary noisy features. Feature selection methods can be classified into two categories: Filter and Wrapper methods [19]. Filter methods are model invariant techniques that use different metrics for selection of features. Wrapper methods on the other hand repeatedly train and retrain the prediction algorithm to determine which feature is the best fit. Generally, Wrapper methods provide better results than Filter methods, but have a huge performance overhead.

Feature importance methods are especially useful in spatial algorithms such as KNN, where the distance of the vectors, and the range of values in each feature drastically influence the prediction results. Methods such as Enhanced Particle Swarm Optimization [13], Gradient Descent performed on the KNN and the like try to find the optimal feature weights. The proposed method tries to find the optimal weights with the help of out-of-bag errors generated while training a Random Forest Model.

Decision Trees are learning algorithms which use splits generated with the help of different metrics to make predictions. Methods such as ID3[20], C4.5[21] and CART[22] are used to make the splits in the decision trees. Ensembling is a technique which uses the collective decision power of multiple learning algorithms combined. It uses the predictions made by multiple leaning algorithms to predict results with a greater accuracy. To perform ensembling, Bootstrap sampling[23] is used to obtain different subsets of data to induce greater diversity in the learning models, and as a result increase the accuracy of predictions. Random Forest Method[24] makes use of Random subspace[25] method which randomly selects features in addition to the selection of a subset of training samples for training the learning algorithm.

Out-of-bag[26] is calculated by checking the error in cases where a particular feature is not taken in account for predictions. This value is easily checkable in models like Random Forest where different algorithms are trained on different subsets of data and features. The Out-of-bag error directly correlates to the importance of that specific feature.

The Feature Importance based Dynamic Scaling method (FIDS method) tries to use the out-of-bag error generated from training the random forest method, to assign individual feature weight for each feature to increase the accuracy of the predictions made by the KNN model.

## Proposed Method

The Feature Importance based Dynamic Scaling method uses feature importance generated from training random forest method on the training data to weigh the features for training the K-Nearest Neighbors Algorithm. K-Nearest Neighbor is especially useful in cases where is a possibility of co-variate shift, and for prediction problems such as item recommendation.

The Z-score normalized feature importance formula in the Random Forest method is given by

$$Error(i,j) = Error\ with\ the\ feature(i,j) - Error\ without\ the\ feature(i,j)$$

$$Average\ Error(i) = \frac{Sum\ of\ all\ trees\ with\ the\ error(i,j)}{number\ of\ trees\ (sum\ of\ all\ j)}$$

We Z-score normalize the features in the data, and then multiply the respective average error for each feature

$$Feature\ Importance(i) = Average\ Error(i)$$

$$Weighted\ Feature\ (i) = Feature\ Importance\ (i) \times Feature(i)$$

This weighted data is used to train the K-Nearest Neighbors algorithms to achieve better results.

## Experimental Results and Discussion

The algorithm was tested on multiple benchmark datasets procured from the University of California Irvine Repository [27] such as Abalone Dataset [28], Balance Scale Dataset[29], Breast Cancer Dataset[30], Cover Type Dataset[31], Income Dataset[32] and Iris Dataset[33].

The experiment was performed in a commercial off the shelf DELL Portable Laptop with 16 Gigabyte DDR4 RAM and Intel 7th Generation 7700HQ series CPU. The experiment environment consisted of UBUNTU 16.04 LTS Operating System, Anaconda Package Management Environment and IPython Notebooks run on a JUPYTER notebook server. The code for the experiment was executed on a Python version 2.7 Interpreter, with the packages being used mainly including Numpy[34], Sci-kit Learn[35], Pandas[36] and Scipy[37].

The dataset procured from the University of California Irvine Repository was first cleaned with a missing value imputation strategy named mean value substitution, which meant that the mean values of particular columns were imputed when any missing value were encountered in the column.

Further, to increase the accuracy of the predictions made by the K-Nearest Neighbors algorithm, as well as to establish a benchmark solution from which an alternate solution could be compared, the dataset was scaled uniformly with a Z-score normalization strategy. The Z-score normalization strategy ensures that each column has Zero Mean and Unit Variance.

After the benchmark solution is set, a Random Forest Model based on multiple decision trees is trained on the dataset to obtained feature importance calculated from out-of-bag errors. This obtained feature importance is individually multiplied to the correspoding columns.

The K-Nearest Neighbors algorithm is trained with the newly scaled data and the accuracy obtained is compared with the results obtained from the benchmark solution procured from running the K-Nearest Neighbors algorithm on the uniformly scaled data.

Table 1. Results for K-Nearest Neighbors Algorithm with K=5

| Sno. | DataSet | Z-Score Scaled | FIDS |
|---|---|---|---|
| 1 | Abalone | 0.52059387 | **0.521312261** |
| 2 | Balance Scale | 0.77724359 | **0.778846154** |
| 3 | Breast Cancer | **0.964850615** | 0.957820738 |
| 4 | Cover Type | 0.6236 | **0.6308** |
| 5 | Income | 0.811885821 | **0.828146935** |
| 6 | Iris | 0.953333333 | **0.96** |

Table 1. showcases and compares the results obtained by running the K-Nearest Neighbor algorithm on various benchmark datasets after performing scaling with the help of Z-Score normalization on the features, and the case where scaling is performed with the help of Feature Importance based Dynamic Scaling method. It can be easily observed that when run on benchmark datasets, the approach with the FIDS algorithm performs comparatively better than a K-Nearest Neighbors algorithm trained on a uniformly scaled dataset.

## Conclusion & Future Work

The Feature Importance based Dynamic Scaling algorithm for assigning feature weights to the individual features of the dataset performs better than assigning uniform weights to the features, as evident in Table 1.

The KNN algorithm is a very rigid algorithm i.e all the data points are considered for every iteration. Artificially infused Randomization in the Random Subspace method helps calculate the Feature importance which forms the basis of the proposed approach. Repeated iterations on randomly selected points helps in figuring out the important features by calculating the out-of-bag error.

This approach is highly parallelizable since constructing multiple decision trees can be parallelized efficiently, meaning the Random Forest Method is highly parallelizable hence the FIDS method is also highly parallelizable.

The feature weights can further be optimized by combining algorithms such as Genetic Algorithms[45] etc. with the FIDS approach, wherein the feature weights obtained from FIDS can be used as the starting seed to converge to a good result for a majority of the cases.

Feature Importance based Dynamic Scaling algorithm can be further extended by using Knowledge-Discovery trees to calculate the nearest neighbors for data points. K-D trees[38], short for Knowledge Discovery trees, are binary trees used to store data points present in k-dimension space. They have been shown to improve the performance [38] of the KNN algorithm.

Accuracy of the predictions made can also be increased by inculcating an additional pre-processing step such as Principal Component Analysis [16]. Principal Component Analysis, generally abbreviated as PCA, projects the data points in a hyper-space where the covariance between the columns is minimized, effectively meaning that inter-relation and redundancy in the data is reduced and the relevant features are extracted. The dimensions with relatively lower variance can be removed to increase the performance of the prediction algorithm, resulting in faster predictions.

Hence, the FIDS method for dynamically scaling the features of a dataset on the basis of feature importance weights obtained from training a Random Forest Model is discussed in detail, and the experimental results obtained are compared with an existing benchmark solution to empirically prove the superiority of the FIDS Method versus a uniformly scaled variant, further a future direction in the form of using the feature weights as seed points for better convergence of the feature weights with the help of combinatorial approach is showcased.